\pdfoutput=1

\documentclass[11pt]{article}

\usepackage[preprint]{acl}

\usepackage{times}
\usepackage{latexsym}

\usepackage[T1]{fontenc}

\usepackage[utf8]{inputenc}

\usepackage{microtype}

\usepackage{inconsolata}
\usepackage{url}
\usepackage{xltabular}
\usepackage{xtab,booktabs}
\usepackage{tabularx}
\usepackage{tabularray}
\usepackage{graphicx}
\usepackage{subcaption}
\usepackage{caption}
\usepackage{amsmath}
\usepackage{amssymb}
\usepackage{mathtools}
\usepackage{amsthm}
\usepackage{multirow}
\usepackage{makecell}
\usepackage{graphicx}
\usepackage{enumitem}
\title{PhyBench: A Physical Commonsense Benchmark for Evaluating Text-to-Image Models}

\author{
Fanqing Meng$^{2,1}$, Wenqi Shao$^{1\dagger}$, Lixin Luo$^{6}$, Yahong Wang$^{3}$, Yiran Chen$^{1}$ \\ Quanfeng Lu$^{5,1}$, Yue Yang$^{2,1}$, Tianshuo Yang$^{1}$, Kaipeng Zhang$^{1}$, Yu Qiao$^{1}$, Ping Luo$^{1,4\dagger}$\\\\
$^{1}$OpenGVLab, Shanghai AI Laboratory \quad $^2$Shanghai Jiao Tong University  \quad $^3$Tongji University \\
$^{4}$The University of Hong Kong \quad $^5$Nanjing University \quad $^6$University of Michigan
}

\begin{document}
\maketitle

\renewcommand{\thefootnote}{\fnsymbol{footnote}}
{\let\thefootnote\relax\footnotetext{
\noindent \hspace{-5mm}$\dagger$ Corresponding Authors: shaowenqi@pjlab.org.cn; pluo@cs.hku.hk 
}   }

\begin{abstract}
Text-to-image (T2I) models have made substantial progress in generating images from textual prompts. However, they frequently fail to produce images consistent with physical commonsense, a vital capability for applications in world simulation and everyday tasks. Current T2I evaluation benchmarks focus on metrics such as accuracy, bias, and safety, neglecting the evaluation of models' internal knowledge, particularly physical commonsense. To address this issue, we introduce PhyBench, a comprehensive T2I evaluation dataset comprising $700$ prompts across four primary categories: mechanics, optics, thermodynamics, and material properties, encompassing $31$ distinct physical scenarios. We assess $6$ prominent T2I models, including proprietary models DALLE3 and Gemini, and demonstrate that incorporating physical principles into prompts enhances the models' ability to generate physically accurate images. Our findings reveal that: (1) even advanced models frequently err in various physical scenarios, except for optics; (2) GPT-4o, with item-specific scoring instructions, effectively evaluates the models' understanding of physical commonsense, closely aligning with human assessments; and (3) current T2I models are primarily focused on text-to-image translation, lacking profound reasoning regarding physical commonsense. We advocate for increased attention to the inherent knowledge within T2I models, beyond their utility as mere image generation tools. The data will be available soon.

\end{abstract}

\begin{figure}[tbp]
  \centering
  \includegraphics[width=\linewidth]{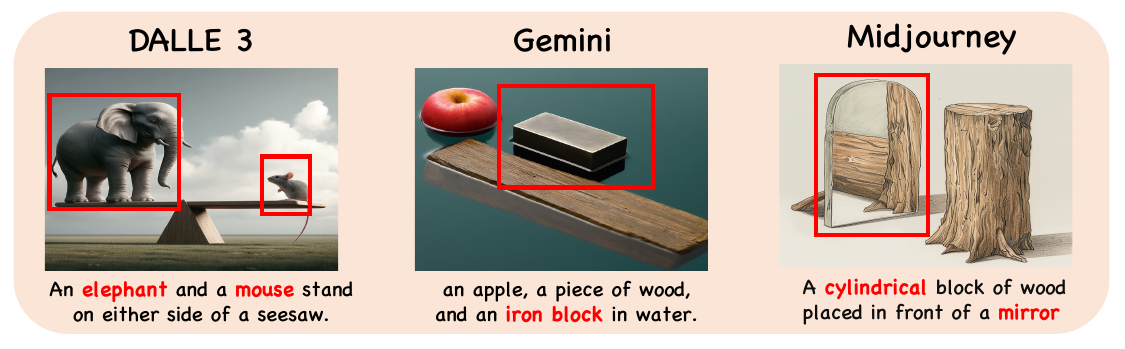}
  \vspace{-15pt}
  \caption{Error images generated by some popular T2I models. }
  \label{fig:error}
  \vspace{-15pt}
\end{figure}

\section{Introduction}

Text-to-image (T2I) models have revolutionized the visualization of abstract concepts and detailed scenes from text descriptions \cite{saharia2022photorealistic,schramowski2023safe,gafni2022make,gal2022image, rombach2022high, betker2023improving}. Beyond generating images from textual prompts, it is crucial for T2I models to adhere to physical commonsense. This adherence ensures the accurate depiction of interactions between objects, light, and shadows, which is essential for constructing realistic world simulators \cite{zhu2024sora}. 
However, it is known that proprietary T2I models often fall short in generating realistic images \cite{borji2023qualitative,farid2022lighting,liu2023hyperhuman}. For example, DALL-E 3 \cite{dalle3} even generates a balanced seesaw with an elephant and a mouse on either side as shown in Fig. \ref{fig:error}, which is physically impossible due to their different masses. 
%
%
%
%

To drive the research in this field, it is crucial to build an evaluation benchmark to measure how much different T2I models grasp the physical commonsense. Such evaluation allows practitioners to identify the strengths and weaknesses of these models, fostering the development of new techniques that better integrate physical commonsense. Prior T2I evaluation benchmarks focus on metrics such as accuracy \cite{huang2023t2i, lee2024holistic}, safety \cite{schramowski2023safe}, and privacy \cite{yang2024position}, as shown in Tab. \ref{tab:benchmarks}. These benchmarks primarily address basic text-to-image translation and do not delve into the physical commonsense behind the image. These benchmarks primarily address basic text-to-image translation and do not explore the underlying physical commonsense in the images. While some studies have highlighted the challenges T2I models face in generating realistic scenes \cite{borji2023qualitative}, a comprehensive quantitative investigation into the physical commonsense of T2I models remains largely unexplored.

Constructing such a benchmark presents several challenges. First, the prompt should be implicit without revealing the underlying physical knowledge, to assess whether T2I models can independently discern physical commonsense. Second, the scenarios showcasing the physical commonsense should be simple enough such that the image depicting user prompts can be generated for T2I models. In this way, the evaluation concentrates on the physical knowledge behind images rather than the alignment between the image and the prompt. Lastly, the prompts in the evaluation set should be comprehensive enough to cover a wide range of physical commonsense to ensure a thorough evaluation of the models' grasp of physical commonsense in different contexts.

\begin{figure*}[htbp]
  \centering
  \scalebox{0.9}{
  \includegraphics[width=\linewidth]{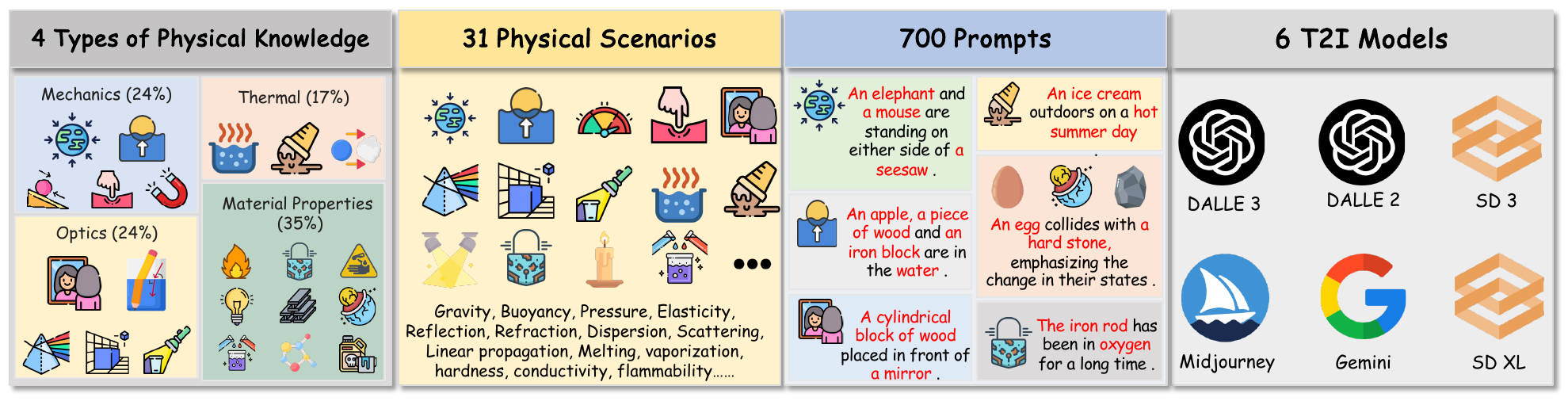}
  }
  \caption{An overview of the proposed PhyBench. It encompasses $4$ types of general physical knowledge spanning $31$ detailed physical scenarios, and $700$ validated prompts, which has been evaluated on $6$ T2I models. }
  \label{fig:phybench}
  \vspace{-10pt}
\end{figure*}

To address these challenges, we introduce \textbf{PhyBench}, focusing on $4$ types of physical knowledge, including optics, mechanics, thermodynamics, and material properties. PhyBench encompasses a total of $700$ prompts, spaning $31$ physical scenarios about gravity, light scattering, ice melt, and more, as shown in Fig. \ref{fig:phybench}. We evaluate $6$ proprietary and open T2I models including DALLE 2 \& 3 \cite{dalle3}, SD XL \cite{podell2023sdxl} \& 3 \cite{sd3}, Midjourney \cite{Midjourney}, and Gemini \cite{team2023gemini}.

\begin{figure*}[htbp]
  \centering
  \scalebox{0.9}{
  \includegraphics[width=\linewidth]{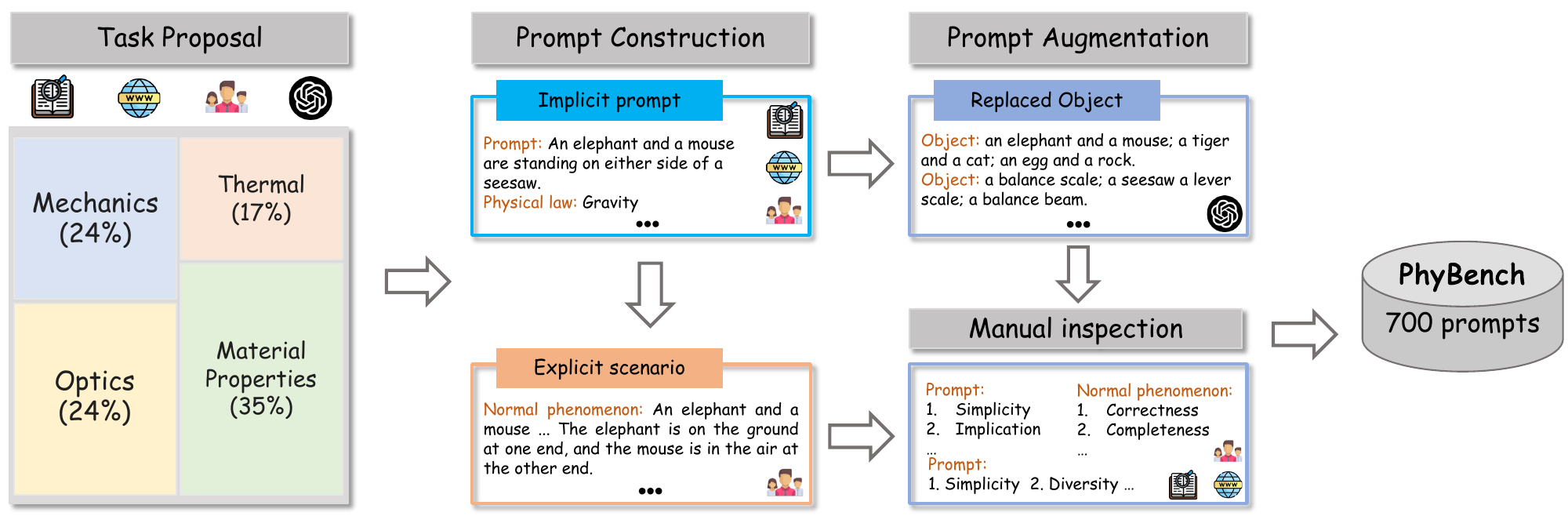}
  }
  \caption{Data collection pipeline of PhyBench. PhyBench comprises $4$ categories of general physical knowledge. For each category, we first consider the detailed physical laws that can be represented by T2I models. Using textbooks, the internet, and other sources, we manually construct appropriate and simple scenarios for each physical law, including the prompt and explicit scenario. Subsequently, we use GPT-4 to expand the prompt by replacing key objects. After rigorous quality checks and filtering, PhyBench includes 700 validated samples.}
  \label{fig:overall}
  \vspace{-10pt}
\end{figure*}

As illustrated in Fig. \ref{fig:overall}, the prompt collection process adheres to a stringent pipeline to cover as many physical commonsense as possible.  On the one hand, the physical commonsense is characterized by $4$ types of physical knowledge as mentioned above. Prompts in optics, mechanics, and thermodynamics evaluate the model’s comprehension of interactions between objects, while prompts in materials assess the model's understanding of the basic attributes of individual objects. On the other hand, we refine each physical scenario with fine-grained physical principles with the help of textbooks and GPT4. Finally, the physical scenarios are specified by various prompts created with GPT4. When creating prompts, we also ensure that the prompts used for drawing are implicit, such as \textit{"an elephant and a mouse standing on either side of a seesaw"}, rather than explicitly stating their positions like explicit prompts. With such a top-down pipeline, PhyBench is curated to comprise $700$ prompts with their physical scenarios recorded.

To accurately evaluate the model's performance on PhyBench, we propose \textbf{PhyEvaler} which leverages GPT-4o \cite{2023GPT4VisionSC} to score the generated images with an item-specific scoring instruction. We show that PhyEvaler aligns well with human rating. The evaluation results reveal that T2I models face great challenges in generating images conforming to physical commonsense.  We summarized several findings as follows.
\begin{itemize}
    \item Current text-to-image (T2I) models often struggle to generate images that adhere to physical commonsense, with the exception of optical scenarios.
    \item Advanced open-source models, such as Stable Diffusion 3 and Stable Diffusion XL, have a significant gap in understanding physical commonsense compared to advanced closed-source models like DALLE 3.
    \item Rewriting implicit prompts by explicitly revealing physical commonsense can significantly improve the accuracy of generated images. This suggests that current T2I models can follow text instructions well but fail to reason about the physical commonsense in image generation.
\end{itemize}




The contributions of our work are three-fold.  i) We build a new evaluation benchmark for text-to-image generation called PhyBench to measure the understanding of physical knowledge for T2I models. ii)
We evaluate various publicly available T2I models on PhyBench, revealing that current T2I models including DALLE 3, Midjourney, and Stable diffusion 3 achieve plain performance in understanding simple physical commonsense. iii) We present an automated evaluation framework called PhyEvaler, which aligns closely with human evaluation results. We hope PhyBench will inspire the community to focus on physical knowledge in image generation, rather than simply treating them as text-to-image translation tools, bringing us closer to a real-world simulator.

\begin{table}[!t]
\caption{Comparison with previous text-to-image benchmarks. Unlike prior datasets, PhyBench quantitatively benchmarks the performance of T2I models in following physical laws.  }
\label{tab:benchmarks}
\centering
\resizebox{\linewidth}{!}{
\begin{tabular}{lccc}
    \toprule
    \textbf{Benchmark} & \textbf{Prompt Type}  & \textbf{Evaluation}  &\textbf{Human Rating}\\
    \midrule
    
    Stable Bias \cite{seshadri2023bias} & Explicit & Gender \& Racial  &  No   \\
    Safe Diffusion \cite{schramowski2023safe} & Explicit & Violence     &  No    \\
    HEIM \cite{lee2024holistic} & Explicit & Align \& Quality  &  Yes    \\
    T2I-CompBench \cite{huang2023t2i} & Explicit  &  Composition  &  Yes    \\
    ImplictBench \cite{yang2024position} & Implicit & Safety     &   No  \\
    \midrule
    \textbf{PhyBench (ours)} & Implicit & Physics knowledge   &    Yes  \\
    \bottomrule
\end{tabular}
}
\vspace{-15pt}
\end{table}

\begin{figure*}[h]
  \centering
  \scalebox{0.9}{
  \includegraphics[width=\linewidth]{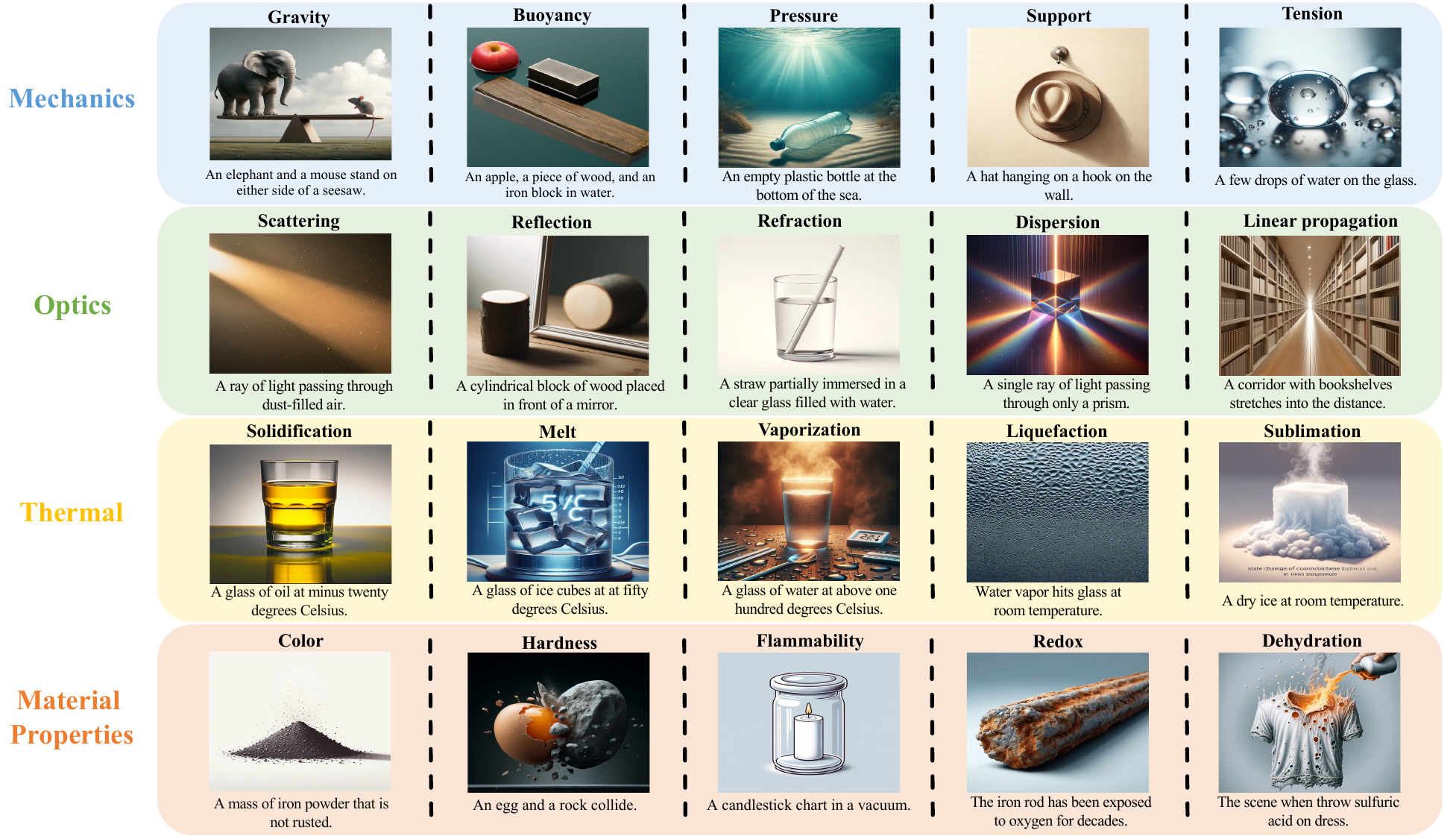}
  }
  \caption{Samples of images induced by our PhyBench in $4$ different aspects by DALLE 3 and Gemini. The bold text represents the detailed physical commonsense and the colored bold text indicates the general physical knowledge.}
  \label{fig:samples}
  \vspace{-10pt}
\end{figure*}

\section{Related work}

\subsection{Text-to-image generation}
Text-to-image (T2I) generation is an evolving field dedicated to creating images from textual descriptions. Initially, research in this area predominantly focused on leveraging Generative Adversarial Networks (GANs) \cite{goodfellow2014generative} and Autoregressive Transformers \cite{vaswani2017attention} to accomplish this task. These early approaches laid the groundwork for generating coherent images based on textual input.
In recent years, diffusion models have emerged as a powerful alternative for T2I generation. These models start with random noise and iteratively refine the image through a de-noising process. Notable examples include the Stable Diffusion series \cite{rombach2022high}, which have set new benchmarks in the field. These models frequently use a frozen text encoder, such as CLIP \cite{radford2021learning}, to transform prompts into embeddings that drive the de-noising process, thereby generating images closely related to the prompts.
Despite these advancements, current state-of-the-art T2I models like DALL-E 3 and Midjourney still face challenges in accurately depicting scenes involving extremely simple physical scenarios. To assess the capabilities of T2I models in generating images that adhere to physical laws, we select and evaluate four state-of-the-art closed-source models: DALL-E 2\cite{betker2023improving}, DALL-E 3 \cite{dalle3}, Midjourney \cite{Midjourney}, and Gemini \cite{team2023gemini}, along with two open-source models: Stable Diffusion XL \cite{rombach2022high} and Stable diffusion 3 \cite{sd3}, using our proposed PhyBench.

\subsection{Benchmarks for text-to-image generation}
The rapid development of text-to-image (T2I) generation models has highlighted the need for comprehensive benchmarks. There are many sophisticated and challenging benchmarks have been introduced to better assess their capabilities. Examples include DrawBench \cite{saharia2022photorealistic} and HE-T2I \cite{petsiuk2022human}, which use hundreds of prompts to evaluate skills in counting, shapes, and writing. Additionally, more complex evaluation tasks, such as those proposed by T2I-CompBench \cite{huang2023t2i}, offer benchmarks for challenging compositional generation, where prompts combine different attributes.
And some research has also emphasized the societal implications of T2I models, particularly concerning safety \cite{schramowski2023safe} and bias \cite{seshadri2023bias}. These previous efforts typically focus on using explicit prompts that clearly specify the content to be included in the image, along with precise positioning and other relevant auxiliary information, to evaluate the model's performance. In contrast, the position \cite{yang2024position} benchmark proposes evaluating models with implicit prompts, which can better assess whether a model possesses the required knowledge.
To this end, we introduce PhyBench, a benchmark that uses implicit prompts to evaluate a model's understanding of physics. By employing implicit prompts, PhyBench aims to provide a more nuanced evaluation of a model's capabilities in generating images based on physical concepts. A detailed comparison of T2I evaluation benchmarks is presented in Tab. \ref{tab:benchmarks}

\section{Our Benchmark: PhyBench}

 Our PhyBench focuses on $4$ main aspects, encompassing $31$ physical scenarios, comprising $700$ prompts, as shown in Fig. \ref{fig:phybench}. We present the details about dataset construction, image generation, and evaluation method in Sec. \ref{sec:dataset-construction}, Sec. \ref{sec:image-gen}, and Sec. \ref{sec:eval}, respectively.

\subsection{PhyBench Construction}\label{sec:dataset-construction}

PhyBench evaluates physical commonsense from four types of physical knowledge as described in the following.

\textbf{Mechanics. } As for mechanics, we categorize it into $6$ aspects according to the type of force: gravity, buoyancy, normal force, elasticity, pressure, and surface tension. We construct physical scenarios that either directly or indirectly reflect these types of forces. For example, for pressure, we designed the implicit prompt \textit{"an empty plastic bottle at the bottom of the sea"}, because the bottle would be compressed under high pressure. Subsequently, we leverage GPT-4 to perform object substitutions (e.g. replacing \textit{"plastic bottle"}) and environment substitutions (e.g. replacing \textit{"at the bottom of the sea"}) to expand the prompts. Ultimately, we have generated \textbf{170} prompts for this aspect.

\textbf{Optics. } In optics, we categorize $6$ aspects based on the physical phenomena of light: reflection, refraction, scattering, dispersion, absorption, and straight-line propagation. Similar to the process in mechanics, we design scenarios for each phenomenon and expand them using GPT-4. For instance, for refraction, due to the different indices of refraction between air and water, we use the prompt \textit{"a plastic straw partially immersed in a clear glass filled with water."} This setup visually demonstrates the refraction by showcasing the apparent discontinuity in the straw. Ultimately, we have generated \textbf{170} prompts for this aspect.

\textbf{Thermodynamics. } The thermodynamic laws that can be depicted by T2I models are limited. Therefore, we choose $6$ types of phase changes to design scenarios and expand prompts. For instance, for sublimation, we design the implicit prompt as \textit{"The state change of dry ice at room temperature"}, where the correct physical phenomenon is the sublimation of dry ice. Ultimately, we have generated \textbf{120} prompts for this aspect.

\textbf{Material Properties. } Physical properties and chemical properties are two aspects of material properties. For physical properties, we categorize them into color, hardness, viscosity, solubility, conductivity, and flame reaction. For hardness, we use implicit prompts involving collisions to reflect the hardness differences between objects. For example, the prompt \textit{"an egg and a hard rock collide"} should produce an image showing that an egg breaks upon colliding with a stone. 

For chemical properties, we consider combustibility, supporting combustion, acidity, redox potential, dehydrating properties, molecular structure, and stability. Since chemical properties often involve chemical reactions, we aim for simplicity and universality in our prompts by avoiding obscure chemical substances and instead using common everyday scenarios. For instance, for redox potential, we use \textit{"The scene after an iron rod has been exposed to oxygen for decades"}, evaluating the correctness of the image by whether it depicts a rusted iron rod. After expanding the prompts, we have generated \textbf{120} prompts for physical properties and \textbf{120} prompts for chemical properties.

\subsection{Image Generation on PhyBench} \label{sec:image-gen}

To investigate the performance of T2I models in response to implicit prompts, we utilize $6$ T2I models detailed in Tab. \ref{tab:models}. Specifically, we produce $4$ images for each implicit prompt using each T2I model to ensure a thorough evaluation and to mitigate the effects of randomness in image generation.

\begin{table}[!t]
\caption{Information of T2I models evaluated on ImplicitBench.}
\label{tab:models}
\centering
\resizebox{\linewidth}{!}{
\begin{tabular}{lccc}
    \toprule
    \textbf{Model} & \textbf{Creator}  & \textbf{\# Params} & \textbf{Access}   \\
    \midrule
    Stable Diffusion 3 \cite{sauer2024fast} & Stability AI  & 2B & Open   \\
    Stable Diffusion XL \cite{podell2023sdxl} & Stability AI  & 3B & Open    \\
    Gemini \cite{team2023gemini} & Google & - & Limited     \\
    Midjourney \cite{Midjourney} & Midjourney & - & Limited     \\
    DALL-E 2 \cite{betker2023improving} & OpenAI & 3.5B & Limited    \\
    DALL-E 3 \cite{dalle3} & OpenAI & - & Limited   \\
    \bottomrule
\end{tabular}
}
\vspace{-15pt}
\end{table}

\subsection{Evaluation Method}
\label{sec:eval}

Our evaluation method aims to score each image generated by text-to-image (T2I) models based on its accuracy in representing correct physical scenarios, ultimately reflecting the rankings and relative differences among different models. For simplicity, the evaluation is divided into two aspects: scene scores \textit{(0, 1, 2)} and physical correctness scores \textit{(0, 1, 2, 3)}. The former assesses whether the image adheres to the prompt by depicting the correct scene, while the latter measures the correctness of the depicted physical principles. For instance, as for the prompt \textit{"an elephant and a mouse are standing on either side of a seesaw"}, if the image includes an elephant and a mouse on a seesaw, the scene score would be 2; And if the elephant's end is on the ground and the mouse's end is in the air, the physical correctness score would be 3.

\begin{figure*}[htbp]
  \centering
  \scalebox{0.9}{
  \includegraphics[width=\linewidth]{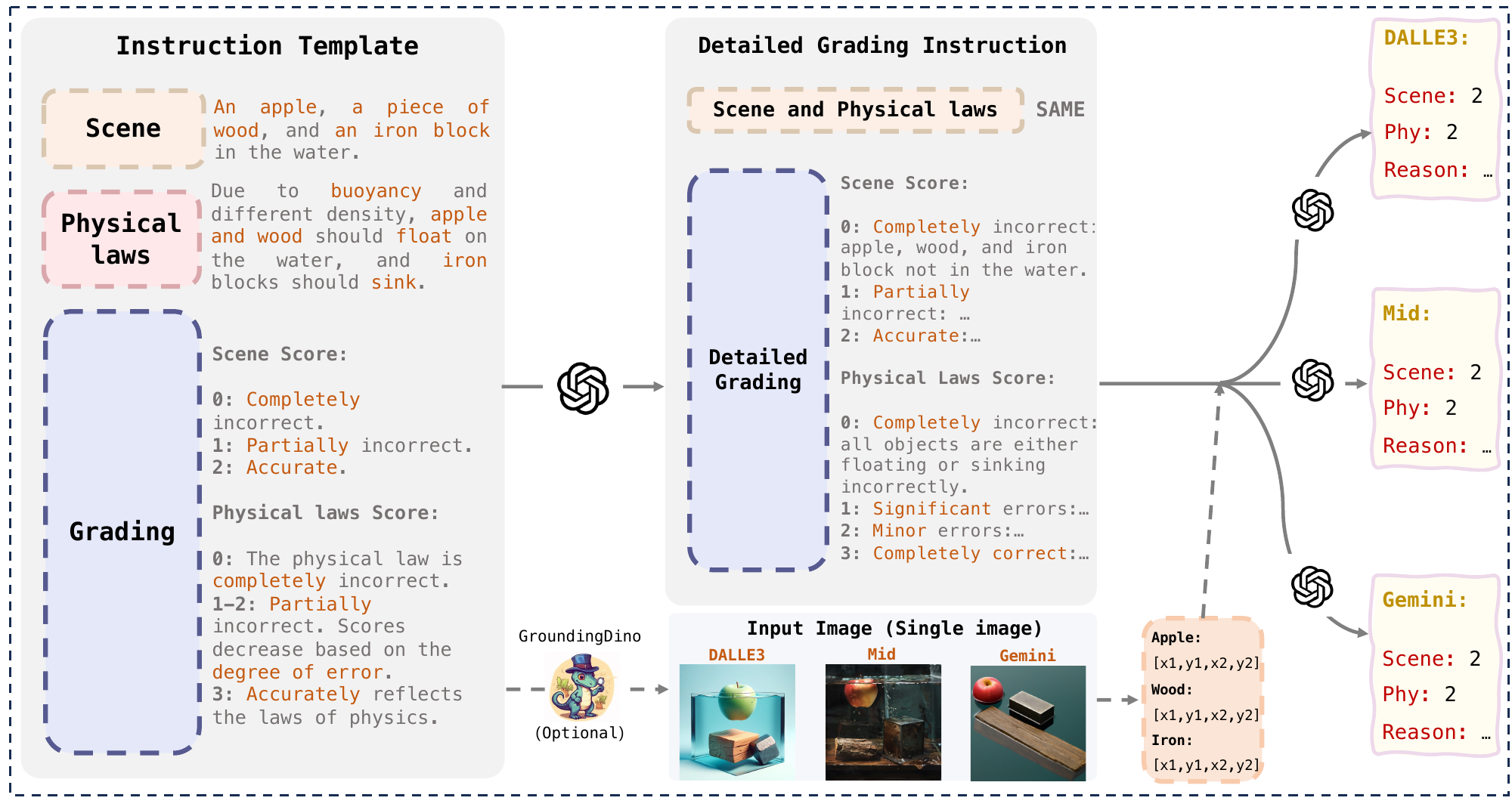}
  }
  \caption{The evaluation method of PhyBench. We pre-establish template scoring criteria and require GPT-4 to generate detailed scoring standards for each scenario. Finally, we use these detailed criteria to score each image. Additionally, for scenarios involving spatial relationship assessments, we provide bounding boxes generated by GroundingDino to enhance the accuracy of machine scoring.}
  \label{fig:eva}
  \vspace{-5pt}
\end{figure*}

\textbf{GPT-4o as Human-Aligned Scorer. } Previous studies have demonstrated that GPT-4V is an effective human-aligned evaluator \cite{wu2024gpt}. Therefore, we employ the more advanced GPT-4o and design special prompts to score images. As illustrated in Fig. \ref{fig:eva}, thanks to GPT-4o's excellent capabilities in understanding pure text and physical laws, we leverage GPT-4o to generate scene-specific scoring criteria for each prompt in our PhyBench dataset. Specifically, we design a grading instruction template that includes descriptions of the current scene and relevant physical laws, along with detailed grading criteria that need to be adapted for each scene. We then prompt GPT-4o to complete the template for each scene. When running the benchmark, we require GPT-4o to score different images according to these detailed criteria and provide justifications for each score. This method of scoring through specific evaluation points has been shown to achieve favorable outcomes in previous work \cite{wu2024gpt}. Our experiments also reveal that this framework align closely with human evaluation results.

\textbf{Improving Spatial Relationship Judgment. } Numerous studies show that multimodal large models are prone to hallucinations \cite{xu2023lvlm,ying2024mmt}. Fortunately, the prompts we use are exceptionally simple, resulting in images that are straightforward and clear with fewer objects. Hence, in our experiments, we observe that GPT-4o exhibits minimal hallucinations in most scenarios. Nevertheless, GPT-4o remains prone to hallucinations in some scenarios involving spatial relationships, such as those related to gravity or buoyancy. To address this issue, we utilize GroundingDino \cite{ren2024grounding} to annotate the bounding boxes of the main subjects in the images and incorporate these annotations into the final scoring prompt. Our experimental results demonstrate that this simple approach is effective in mitigating the erroneous scoring induced by GPT-4o's hallucinations.

\section{Experiments}
\vspace{-3pt}

In this section, We quantitatively assess the ability of different T2I models to understand physical commonsense on our PhyBench. 



\vspace{-3pt}
\subsection{Experiments Setup}


\textbf{Evaluation Metric. } We provide the average scores for scene and physical correctness assessments by GPT-4o, alongside the average results of human scoring. To better measure the alignment between machine scoring and human scoring, we calculate the Pearson correlation coefficient based on the overall scores from both machine and human evaluations for each sample and report the average number. A detailed description is in Sec. \ref{sec:experiment}

\vspace{-3pt}
\subsection{Quantitative Evaluation}

\begin{figure}[ht]
\begin{center}
\scalebox{0.75}{
\centerline{\includegraphics[width=\columnwidth]{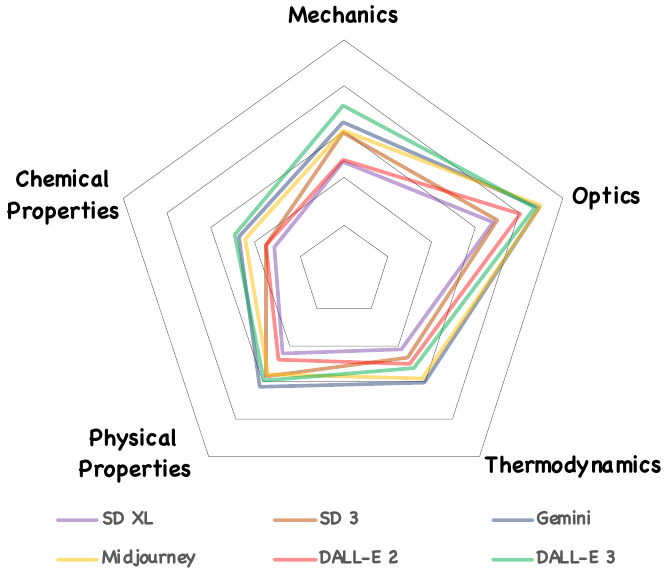}}
}
\caption{The overall scores of different T2I models toward PhyBench across different physical knowledge.}
\label{fig:overall_result}
\vspace{-10pt}
\end{center}
\end{figure}

\textbf{Mechanics. } As shown in Tab. \ref{tab:force}, we find that in the type of mechanics, although DALL-E 3's physics score is only $1.71$, it significantly surpasses other models. However, the physics scores of other models do not even exceed $1.5$, indicating that current models struggle to understand physical scenarios such as gravity and buoyancy accurately.

It is worth noting that in scenarios involving judgment of spatial relationships followed by scoring, GPT-4o sometimes experiences hallucinations. By incorporating the bounding box information of objects as described in Sec. \ref{sec:eval}, we have achieved more accurate machine scoring results. As shown in Tab. \ref{tab:ground}, with the inclusion of bounding box information, the average machine scoring results are closer to human evaluation outcomes, and the correlation between machine scores and human scores significantly improves.

\textbf{Optics. }  As shown in Tab. \ref{tab:light}. In optical scenarios, all models perform well, with Midjourney excelling in both machine and manual scoring, even surpassing a score of $2.5$. We find that for some simple laws, such as light travelling in straight lines and scattering, nearly all models perform well, leading to high overall scores. We believe this is because optical laws are present in pre-trained images, resulting in a better understanding of optics.

\textbf{Thermodynamics. } As shown in Tab. \ref{tab:heat}, we find that all models struggle with thermodynamic scenarios. Even the best-performing model, Gemini, achieves only a $1.31$ score in simulating the physical scenarios of thermodynamics. The scenarios are typically straightforward, often involving a single object, such as a cup of water above $100$ degrees Celsius. However, most current models fail to understand these implicit prompts, either incorrectly depicting the state of objects under specified temperature conditions or misrepresenting them entirely.

\textbf{Material Properties. } Material properties include physical and chemical properties, similar to thermodynamics, generally involving only one object. For example, \textit{"An image of iron powder"} reflects the color of chemicals, and \textit{"A scene of an iron rod after being exposed to oxygen for decades"} reflects redox properties. As shown in Tab. \ref{tab:phy} and Tab. \ref{tab:chem}, current models largely fail to understand such prompts, with the highest score for physical properties being only $1.28$ and the top score for chemical properties just $0.84$. We hypothesize that this may be due to the lack of these specific scenarios in the training data.


\begin{table}[h]
\caption{Specific scores of different T2I models for mechanical laws. }
\vspace{-10pt}
\label{tab:force}
\vskip 0.1in
\begin{sc}
\centering
\resizebox{\linewidth}{!}{
\begin{tabular}{l|c|c|c|c|c|c}
    \toprule
   \multicolumn{1}{c|}{\multirow{2}{*}{\textbf{Model}}}  & \multicolumn{4}{|c|}{\textbf{Score}($\uparrow$)} & \multicolumn{1}{c}{\multirow{2}{*}{\textbf{Corr($\uparrow$)}}} \\
   \cline {2-5}
   ~ &  \textbf{Sce} & \textbf{Sce(hum.)} & \textbf{Phy} & \textbf{Phy(hum.)} & ~ \\
    \midrule
    SD 3   &  1.76   &  1.74   &  1.18  &  1.23   &  0.89 \\
    SD XL   & 1.46   &  1.50 &  0.91 &  0.87   & 0.85 \\
    \midrule
    Gemini       & 1.83 & 1.81 & 1.37 &  1.39  & 0.78 \\
    Midjourney   & 1.82 & 1.77 &  1.33 &  1.24  & 0.76 \\
    DALL-E 2     & 1.53 & 1.49 &  1.15 &  0.92  & 0.72\\
    DALL-E 3   & \textbf{1.88} &\textbf{ 1.87} & \textbf{1.64 } & \textbf{1.71} & 0.77 \\
    \midrule
    SD XL(rewrite)   &  1.62  &  1.69  &  1.23  &  1.20  & 0.90  \\
    DALL-E 3(rewrite)   &  \textbf{1.93 } &  \textbf{1.92}  &  \textbf{2.09}   &  \textbf{2.02}  &  0.84  \\
    \bottomrule
\end{tabular}
}
\end{sc}
\vspace{-0.1in}
\vspace{-11pt}
\end{table}

\begin{table*}[h]
\caption{The effect of adding spatial relationship information in GPT-4o. We only show examples of distinguishing by positional relationship, including some samples in gravity, buoyancy, and object pressure.}
\vspace{-10pt}
\label{tab:ground}
\vskip 0.1in
\begin{sc}
\centering
\resizebox{\textwidth}{!}{
\begin{tabular}{l|c|c|c|c|c|c|c|c|c|}
    \toprule
   \multicolumn{1}{c|}{\multirow{2}{*}{\textbf{Model}}}  & \multicolumn{6}{|c|}{\textbf{Score}($\uparrow$)} & \multicolumn{2}{c}{\textbf{Correlation($\uparrow$)}} \\
   \cline {2-9}
   ~ &  \textbf{Sce} & \textbf{Sce(ground)} & \textbf{Sce(hum.)} & \textbf{Phy} & \textbf{Phy(ground)} & \textbf{Phy(hum.)} & \textbf{w/o ground} & \textbf{w/ ground} \\
    \midrule

    SD 3   &  1.81   &   1.68  &   1.58  &   0.80   &   0.68   &  0.66    &  0.75   &  \textbf{0.86}   \\

    SD XL   & 1.01  & 1.03  & 1.08  & 0.21  & 0.25  &  0.30  &  0.82  & \textbf{0.88}  \\
    \midrule
    Gemini    &  1.77  &  1.71  & 1.66  &   1.05  & 1.14   &  1.14   & 0.67  & \textbf{0.78}    \\
    Midjourney  &  1.79  &  1.65  & 1.56 &  0.92  & 0.82  &  0.74  &  0.69  &  \textbf{0.80}      \\
    DALL-E 2  & 1.31  &  1.23  &  1.10  &  0.87  &   0.65 &  0.57  &  0.59  &  \textbf{0.71}   \\
    DALL-E 3   & 1.97   &  1.97  &  1.87   &   1.36 &  1.56   &  1.56   &  0.70  &  \textbf{0.81}  \\
    \bottomrule
\end{tabular}
}
\vspace{-0.1in}
\end{sc}
\end{table*}

\begin{table}[h]
\caption{Specific scores of different T2I models for optics laws. }
\vspace{-10pt}
\label{tab:light}
\vskip 0.1in
\begin{sc}
\centering
\resizebox{\linewidth}{!}{
\begin{tabular}{l|c|c|c|c|c|c|}
    \toprule
   \multicolumn{1}{c|}{\multirow{2}{*}{\textbf{Model}}}  & \multicolumn{4}{|c|}{\textbf{Score}($\uparrow$)} & \multicolumn{1}{c|}{\multirow{2}{*}{\textbf{Corr($\uparrow$)}}} \\
   \cline {2-5}
   ~ &  \textbf{Sce} & \textbf{Sce(hum.)} & \textbf{Phy} & \textbf{Phy(hum.)} & ~ \\
    \midrule
    SD 3   &  1.72  &   1.74  &    1.77  &   1.75  &  0.90  \\
    SD XL   &  1.66 &  1.64  &  1.92   &    1.78  & 0.89 \\
    \midrule
    Gemini       & 1.92 & 1.94 & 2.58 &  2.51  & 0.73\\
    Midjourney   & 1.94 & 1.95 &  \textbf{2.64} &  \textbf{2.56}  & 0.74 \\
    DALL-E 2     & 1.84 & 1.81 &  2.21 &  2.21  & 0.74\\
    DALL-E 3   & \textbf{1.95} & \textbf{1.95} & 2.53  & 2.41 & 0.73 \\
    \midrule
    SD XL(rewrite)   &  1.79  &  1.77  &  2.09  &  2.13  & 0.88  \\
    DALL-E 3(rewrite)  & \textbf{1.95}  & \textbf{1.96}   &  \textbf{2.67}  &  \textbf{2.65}   &  0.78    \\
    \bottomrule
\end{tabular}
}
\vspace{-0.1in}
\vspace{-10pt}
\end{sc}
\end{table}

\begin{table}[h]
\caption{Specific scores of different T2I models for thermodynamics laws.}
\vspace{-10pt}
\label{tab:heat}
\vskip 0.1in
\begin{sc}
\centering
\resizebox{\linewidth}{!}{
\begin{tabular}{l|c|c|c|c|c|c}
    \toprule
   \multicolumn{1}{c|}{\multirow{2}{*}{\textbf{Model}}}  & \multicolumn{4}{|c|}{\textbf{Score}($\uparrow$)} & \multicolumn{1}{c}{\multirow{2}{*}{\textbf{Corr($\uparrow$)}}} \\
   \cline {2-5}
   ~ &  \textbf{Sce} & \textbf{Sce(hum.)} & \textbf{Phy} & \textbf{Phy(hum.)} & ~ \\
    \midrule
    SD 3   &   1.70   &  1.82    &  0.51    &  0.52   &   0.89  \\
    SD XL   & 1.50  & 1.56  &  0.59  &  0.54 &  0.88 \\
    \midrule
    Gemini       & \textbf{1.68} & \textbf{1.78} &  \textbf{1.45}   &  \textbf{1.31}  &  0.79  \\
    Midjourney   & 1.58 & 1.73 &  1.28   &   1.15  &  0.82 \\
    DALL-E 2     & 1.44 & 1.50 &  1.02   &  0.98   &  0.75  \\
    DALL-E 3   & 1.63 & 1.53 &  1.14   &  1.09   &  0.74   \\
    \midrule
    SD XL(rewrite)   &  1.69  &  1.85  &  1.98  &  2.08  &  0.87 \\
    DALL-E 3(rewrite)   &  \textbf{1.88 } & \textbf{ 1.94}  &  \textbf{2.40}   &  \textbf{2.46 } &  0.80  \\
    \bottomrule
\end{tabular}
}
\vspace{-0.1in}
\vspace{-2pt}
\end{sc}
\end{table}

\begin{table}[h]
\caption{Specific scores of different T2I models for physical properties of material properties.}
\vspace{-10pt}
\label{tab:phy}
\vskip 0.1in
\begin{sc}
\centering
\resizebox{\linewidth}{!}{
\begin{tabular}{l|c|c|c|c|c|c}
    \toprule
   \multicolumn{1}{c|}{\multirow{2}{*}{\textbf{Model}}}  & \multicolumn{4}{|c|}{\textbf{Score}($\uparrow$)} & \multicolumn{1}{c}{\multirow{2}{*}{\textbf{Corr($\uparrow$)}}} \\
   \cline {2-5}
   ~ &  \textbf{Sce} & \textbf{Sce(hum.)} & \textbf{Phy} & \textbf{Phy(hum.)} & ~ \\
    \midrule
    SD 3     &   1.60   &   1.67   &   1.15    &    1.18   &  0.86     \\
    SD XL   & 1.34  & 1.31 & 0.97  & 0.93  & 0.87 \\
    \midrule
    Gemini       & 1.71 & 1.81 & \textbf{1.39} &  \textbf{1.28}  & 0.83 \\
    Midjourney   & 1.63 & 1.80 & 1.13  &  0.99  & 0.81 \\
    DALL-E 2     & 1.40 & 1.44 & 1.04 &  0.94  & 0.65\\
    DALL-E 3   & \textbf{1.77} & \textbf{1.84} & 1.23  & 1.10 & 0.83 \\
    \midrule
    SD XL(rewrite)   &  1.44  &  1.47  &  1.60   &  1.61  & 0.88  \\
    DALL-E 3(rewrite)   & \textbf{1.89 }  &  \textbf{1.88}  &   \textbf{2.42}  & \textbf{ 2.45 } &  0.84   \\
    \bottomrule
\end{tabular}
}
\vspace{-0.1in}
\vspace{-10pt}
\end{sc}
\end{table}

\begin{table}[h]
\caption{Specific scores of different T2I models for chemical properties of material properties.}
\vspace{-10pt}
\label{tab:chem}
\vskip 0.1in
\begin{sc}
\centering
\resizebox{\linewidth}{!}{
\begin{tabular}{l|c|c|c|c|c|c}
    \toprule
   \multicolumn{1}{c|}{\multirow{2}{*}{\textbf{Model}}}  & \multicolumn{4}{|c|}{\textbf{Score}($\uparrow$)} & \multicolumn{1}{c}{\multirow{2}{*}{\textbf{Corr($\uparrow$)}}} \\
   \cline {2-5}
   ~ &  \textbf{Sce} & \textbf{Sce(hum.)} & \textbf{Phy} & \textbf{Phy(hum.)} & ~ \\
    \midrule
    SD 3   & 1.29   & 1.37  & 0.32   &  0.41 & 0.88  \\
    SD XL   & 1.27  & 1.36  & 0.22  &  0.23 & 0.90 \\
    \midrule
    Gemini       & \textbf{1.61} & \textbf{1.68} & 0.62 &  0.71  & 0.70 \\
    Midjourney   & 1.36 & 1.57 & 0.62  &  0.69  & 0.74 \\
    DALL-E 2     & 1.30 & 1.38 & 0.45 &  0.36  & 0.70\\
    DALL-E 3   & 1.58 & 1.64 & \textbf{0.79}  & \textbf{0.84} & 0.83 \\
    \midrule
    SD XL(rewrite)   &  1.43  &  1.58   &  1.06  &  1.11  &  0.89 \\
    DALL-E 3(rewrite)   & \textbf{1.77}   & \textbf{1.70}   &  \textbf{1.82 }  &  \textbf{1.82}  &  0.87  \\
    \bottomrule
\end{tabular}
}
\vspace{-0.1in}
\vspace{-5pt}
\end{sc}

\end{table}

\subsection{Takeway Findings}
Thorugh extensive evaluation, we have several observations. 1) The scenes we design are simple enough that most models are able to depict them based on the prompts. 2) The machine scoring results show a high correlation with manual evaluation outcomes under our previous designs, indicating that GPT-4o can serve as an excellent human-aligned scorer on PhyBench. 3) As shown in Fig. \ref{fig:overall_result}, although the models can render the scenes, their performance on physical correctness is generally poor, with only optical scenarios showing relatively better results. 4) Open-source models exhibit a significant gap in understanding physical commonsense compared to proprietary models. In Fig. \ref{fig:different_models} in appendix, we present qualitative visual results and compare the performance of different models.

\vspace{-3pt}

\section{Discussion}

\vspace{-5pt}

In this section, we conduct robustness analysis experiments, including prompt rewriting and the use of different T2I models. The experimental results indicate that current T2I models primarily possess superficial text-to-image translation functionalities but lack internal knowledge, such as an understanding of physical commonsense. In section .\ref{sec:case}, we also do some detailed case analysis.

\subsection{Ablation Study}

\vspace{-3pt}

\begin{figure}[tbp]

    \vspace{-10pt}
  \centering
  
  \scalebox{0.95}{
  \includegraphics[width=\linewidth]{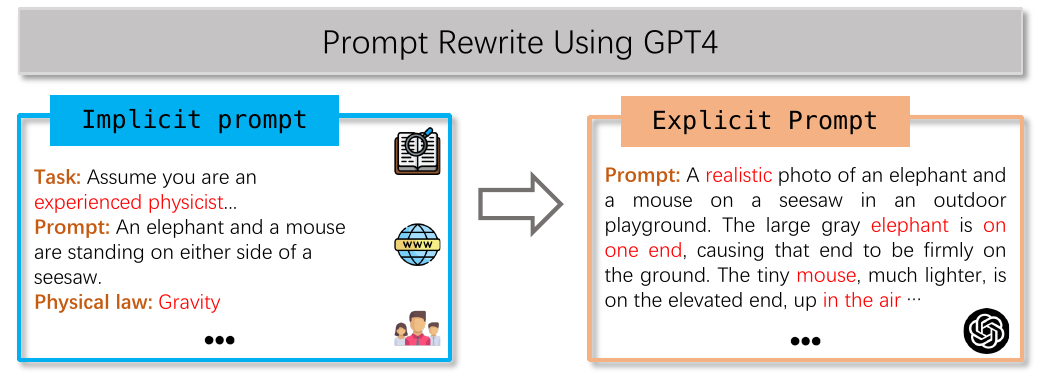} 
  }
  \vspace{-5pt}
  \caption{The PhyRewriter framework. We utilize GPT-4's strong textual capabilities to generate explicit prompts that accurately depict physical phenomena.}
  \label{fig:rewrite}
\end{figure}

\textbf{Rewriting Prompt. } We employ a straightforward strategy to compel models to generate images that comply with physical commonsense, as shown in Fig. \ref{fig:rewrite}. GPT-4's powerful textual capabilities enable it to accurately understand simple physical scenarios. We input the original prompt and the physical laws, along with several in-context examples, to prompt GPT to generate an explicit prompt that includes the outcomes resulting from these physical laws. We have employed explicit prompts corresponding to implicit prompts and selected SDXL and DALL-E 3 as representatives of open-source and proprietary models, respectively. Note that we use \textit{"My prompt has full detail so there is no need to add more:"} as a prefix for DALL-E 3 to prevent it from rewriting the prompt.

As shown in Tab. \ref{tab:heat}, Tab. \ref{tab:phy}, and Tab. \ref{tab:chem}, the results demonstrate that this simple strategy significantly enhances performance for all models. For instance, in thermodynamic scenarios, SDXL with explicit prompts outperforms all models using implicit prompts; after employing explicit prompts, DALL-E 3's score increased from 1.09 to 2.46. This demonstrates that the models can generate images from PhyBench but fail to accurately render images from implicit prompts due to a lack of physical commonsense.


\begin{figure}[tbp]
  \centering
  \scalebox{0.8}{
  \includegraphics[width=\linewidth]{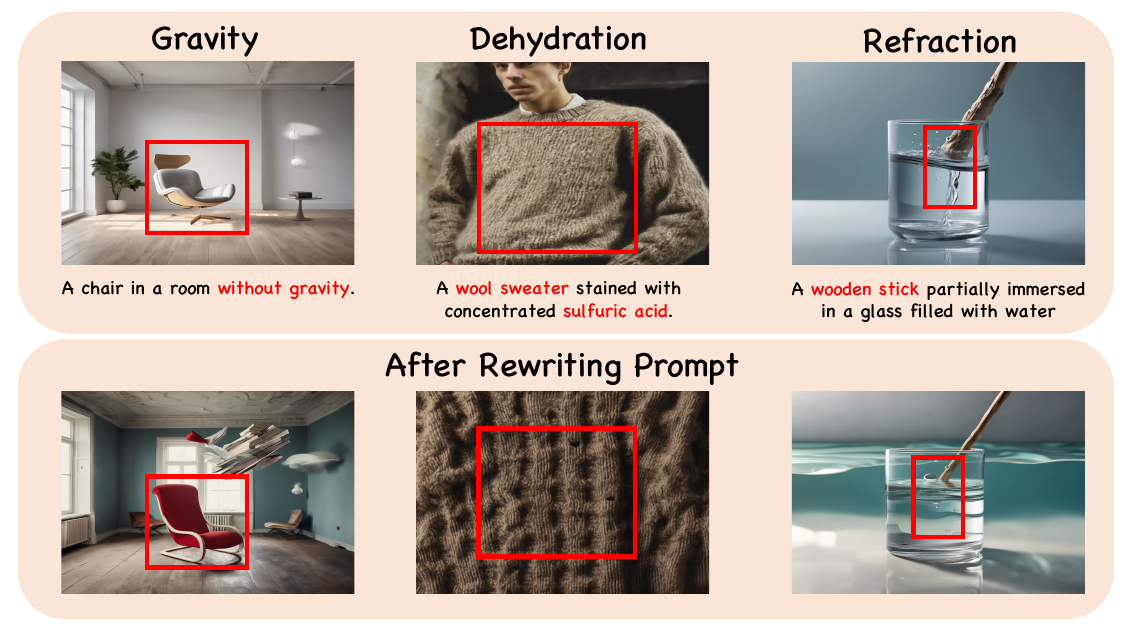}
  }
  \vspace{-5pt}
  \caption{Error cases of RPG in PhyBench.}
  \label{fig:rpg}
  \vspace{-15pt}
\end{figure}

\textbf{Other Text-to-Image models with Reasoning Capabilities. } We also consider T2I models with reasoning capabilities. RPG \cite{yang2024mastering} is a T2I agent framework that leverages the reasoning abilities of LLMs to generate images. We use RPG for experiments in PhyBench. Unfortunately, the success rate of RPG in generating images is notably low, as also reported in their GitHub repository. This limitation hinders quantitative comparison, so we perform a qualitative analysis instead.


Through comparison of numerous images, we find that RPG struggles to accurately model physical scenarios. As shown in Fig. \ref{fig:rpg}, even with simple scenes, RPG fails to generate images correctly, whether using the original or rewritten prompts. We attribute this to two main factors: 1) current T2I models with reasoning capabilities focus on layout and object details (such as texture and shape) but lack an understanding of physical laws. 2) Even with explicit prompts, the models still fail to generate the correct images. We believe this is due to a lack of training data for these types of images.


\vspace{-6pt}
\section{Conclusion}

\vspace{-6pt}

In this paper, we introduce \textbf{PhyBench}, the first comprehensive and quantitative benchmark for evaluating T2I models' understanding of physical commonsense. PhyBench covers $4$ main aspects: mechanics, optics, thermodynamics, and material properties. It includes $31$ specific physical scenarios and $700$ manually reviewed prompts. Besides, we develop \textbf{PhyEvaler}, using GPT-4o for a comprehensive evaluation of six popular T2I models, showing a strong correlation with human evaluation. Our results reveal that current T2I models struggle to generate images that adhere to physical commonsense, indicating that while these models excel in text-to-image translation, they lack in-depth physical knowledge.

\section{Limitations}

Through extensive experiments, we find that current T2I models lack an understanding of fundamental physical commonsense. However, strategies such as prompt rewriting or agent frameworks do not address the core issue. We aim to develop a T2I model capable of autonomously understanding and applying knowledge, which we leave for future work.


\bibliography{custom}

\newpage
\clearpage
\appendix






\section{Experiments} \label{sec:experiment}
\begin{figure}[tbp]
  \centering
  \includegraphics[width=1.0\columnwidth]{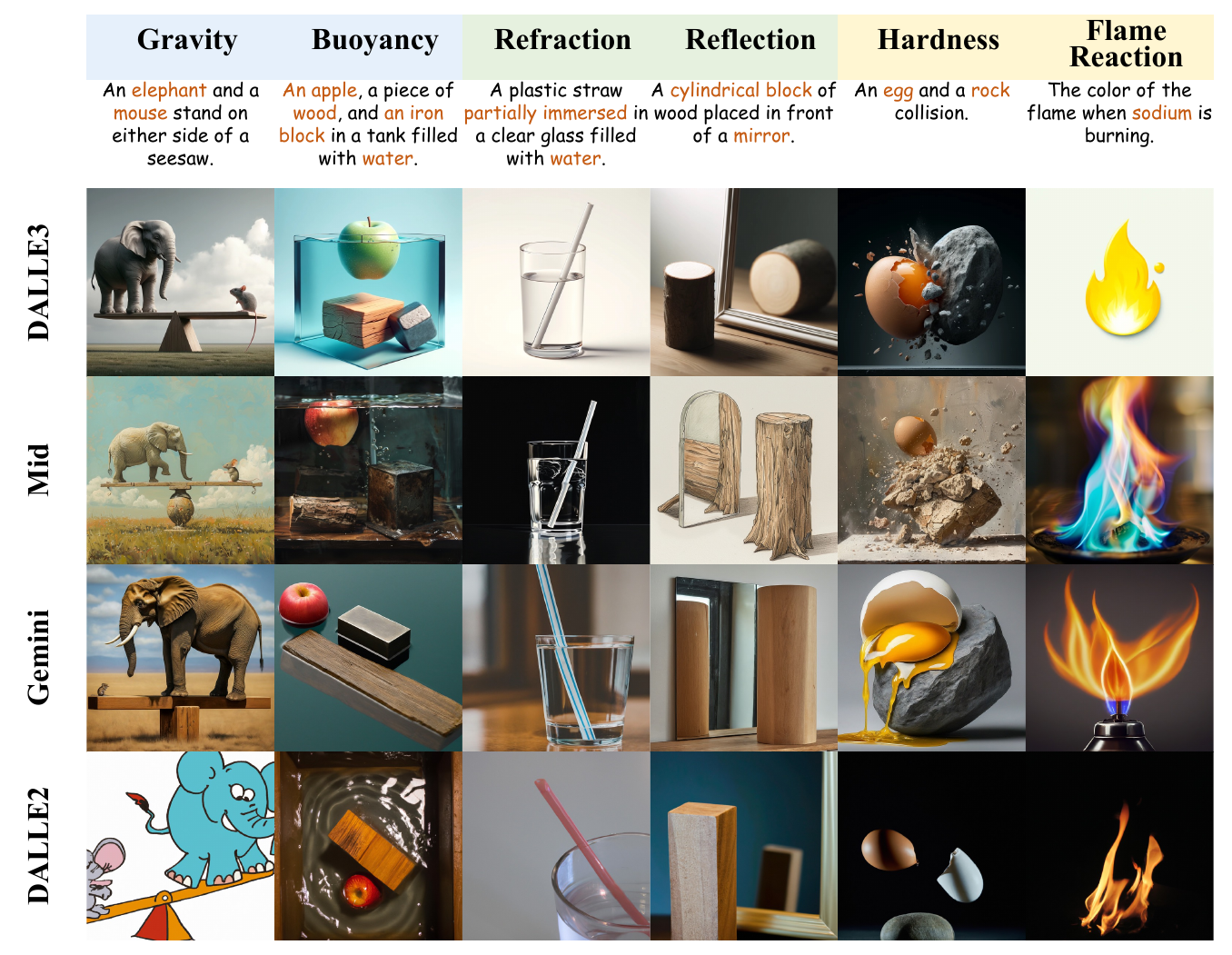}
  \caption{Visual comparisons of different T2I models toward PhyBench across $4$ aspects.}
  \label{fig:different_models}
  \vspace{-8pt}
\end{figure}

In this section, we present qualitative visual results and compare the performance of different models, as shown in Fig. \ref{fig:different_models}. We find that even in the simplest scenarios, current text-to-image (T2I) models are prone to errors. For example, for the prompt \textit{"A cylindrical block of wood placed in front of a mirror"}, DALL-E 3 and Midjourney incorrectly depict objects in a mirror. Similarly, for \textit{"An apple, a piece of wood, and an iron block in a tank filled with water"}, DALL-E 3, Gemini, and Midjourney all misrepresent the state of one or more objects, while DALL-E 2 omits an object entirely. 

\subsection{Experiments Setup}

\textbf{Prompt format. } To ensure fairness, for Midjourney, Gemini, and DALLE 2, we use the original prompt directly. For DALL-E 3 and Gemini, we use the prefix \textit{"I NEED to test how the tool works with extremely simple prompts. DO NOT add any detail, just use it AS-IS:"} to prevent them from rewriting the prompt. For Stable Diffusion models, we employ GPT-4 to reformat the prompt into several short phrases suitable for the models.

\textbf{Human Evaluation. } We require all paper authors to score each image according to the same criteria: 0-2 points for scene accuracy and 0-3 points for adherence to physical laws. Each image receives scores from 10 different human evaluators, and the average of these scores is taken as the final human rating.

\section{Discussion}

\subsection{Case Analysis} \label{sec:case}

\begin{figure}[tbp]
  \centering
  \includegraphics[width=\linewidth]{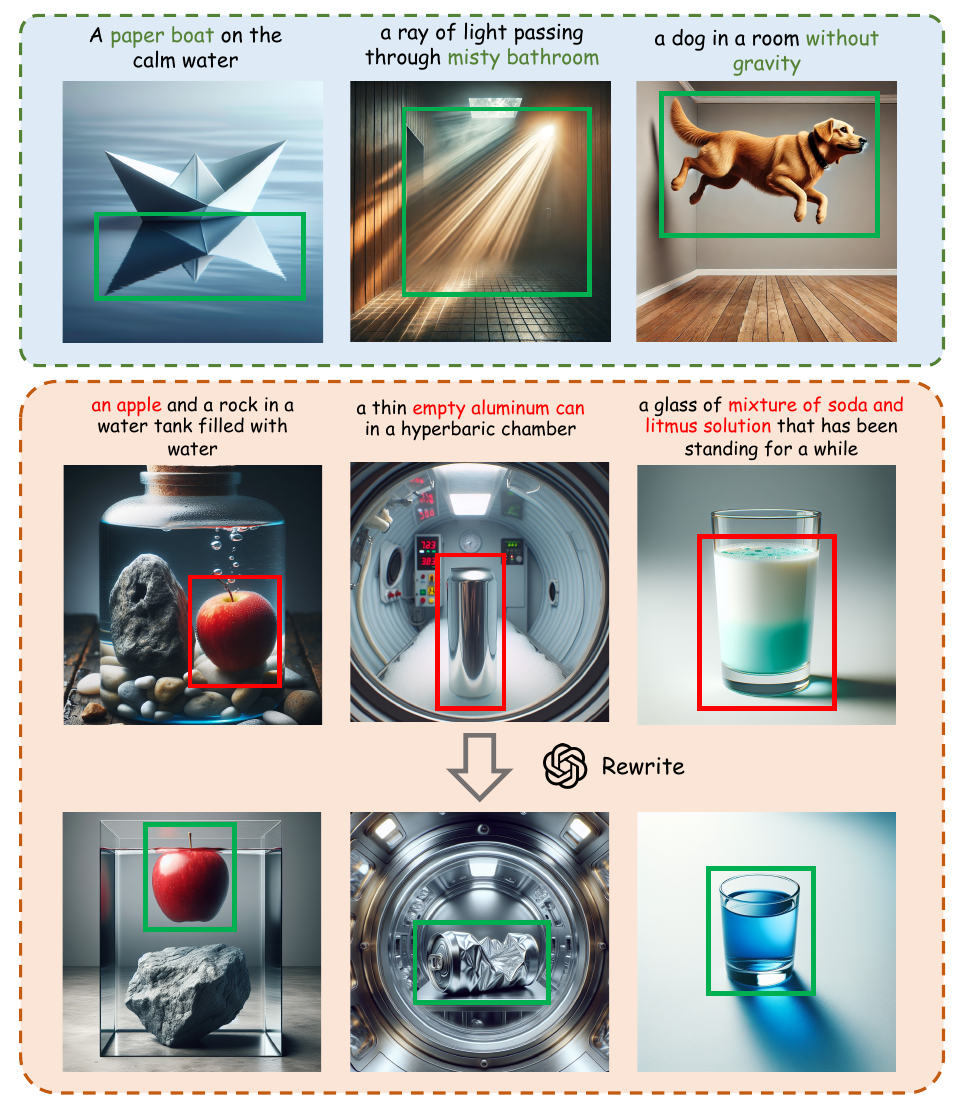}
  \vspace{-15pt}
  \caption{Case analysis of images generated by DALL-E 3. The three examples in the upper row are successful cases where DALL-E 3 is able to generate physically correct images. The three examples in the lower row demonstrate failure cases, with erroneous sections highlighted in red boxes. Following prompt rewriting, the model generates physically correct images, with corrected sections marked in green boxes.}
  \label{fig:case-study}

  \vspace{-15pt}
\end{figure}

As shown in Fig. \ref{fig:case-study}, the three images in the upper row are generated correctly, all following the physical laws (reflection, scattering of light, and weightlessness). These physical laws are more widely reflected in natural images, allowing image generation models to learn and replicate them effectively.

However, for more challenging scenes in our dataset, these models fail to accurately represent physical laws in the generated images. Fig. \ref{fig:case-study} shows three failure cases. For the scene \textit{"an apple and a rock in a water tank filled with water"}, based on the law of buoyancy, the apple, being less dense than water, would float on the surface, and the rock, being denser, would sink to the bottom. However, the generated image depicts both the apple and the rock at the bottom of the tank. GPT-4 rewrites the prompt to includes this information: \textit{"A realistic image of an apple floating on the surface of clear water in a glass water tank. At the bottom of the tank, a rock is clearly visible resting on the tank floor. The water tank is transparent to clearly show the apple floating above and the rock sunk below"}. Using the revised prompt, DALL-E 3 is able to generate an image showing the apple floating and the rock at the bottom.

Similarly, the scene "a thin empty plastic bottle in a hyperbaric chamber" presents another challenge. Given the high atmospheric pressure inside such chambers, an empty, thin aluminum can will crumple or implode due to the external pressure being far greater than the internal pressure. When prompted with the basic prompt, DALL-E generates a smooth, glossy can in a hyperbaric chamber, which is a simple composition of the elements in the prompt. After refining the prompt to "\textit{A realistic scene inside a hyperbaric chamber showing a crumpled thin aluminum can due to the high external pressure}", the model successfully depicted the can's deformation.

For material properties, the scene \textit{"a glass of mixture of baking soda and litmus solution that has been standing for a while"} involves the knowledge of acidity and alkalinity. As an alkaline substance, when mixed with a litmus solution, it will turn the litmus blue. With the original prompt, DALL-E generates a glass of liquid with an incorrect milky white and green colors. When given the revised prompt \textit{"A realistic close-up of a clear glass filled with a blue-colored liquid. ..."}, the model is able to generate the correct blue color.

\end{document}